\documentclass[10pt]{article} % For LaTeX2e
\usepackage[preprint]{tmlr}

\usepackage{amsmath,amsfonts,bm}

\def\eqref#1{equation~\ref{#1}}
\def\1{\bm{1}}

\DeclareMathAlphabet{\mathsfit}{\encodingdefault}{\sfdefault}{m}{sl}
\SetMathAlphabet{\mathsfit}{bold}{\encodingdefault}{\sfdefault}{bx}{n}

\usepackage{float}
\usepackage{hyperref}
\usepackage{url}
\usepackage{booktabs}
\usepackage{tabularx}
\usepackage{needspace}
\usepackage{multirow}
\usepackage{amsmath}

\title{Reproducing INP-Former: Reconstructing Intrinsic Normal Prototypes for Industrial Anomaly Detection}

\author{\name Yunfei Liang \email yl84070@student.uni-lj.si \\
      \addr Faculty of Computer and Information Science\\
      University of Ljubljana
      \AND
      \name Juan Pablo Osorio Rivera \email jo56909@student.uni-lj.si \\
      \addr Faculty of Computer and Information Science\\
      University of Ljubljana}

\def\month{05}  % Insert correct month for camera-ready version
\def\year{2026} % Insert correct year for camera-ready version
\def\openreview{\url{https://openreview.net/forum?id=XXXX}} % Insert correct link to OpenReview for camera-ready version

\usepackage{amssymb}
\usepackage[section]{placeins}
\usepackage[most]{tcolorbox}
\definecolor{EditGold}{HTML}{AE7814}
\definecolor{EditCream}{HTML}{FFF6D8}
\definecolor{ReviewRed}{HTML}{AC4848}
\definecolor{ReviewPink}{HTML}{FFF3F1}
\newtcolorbox{editnote}[1]{enhanced,breakable,colback=EditCream,colframe=EditGold,
 boxrule=0.6pt,arc=1mm,left=2mm,right=2mm,top=1mm,bottom=1mm,
 title={[FILL IN HERE] #1},fonttitle=\small\bfseries,fontupper=\small,
 before skip=7pt,after skip=8pt}
\newtcolorbox{reviewpassage}[1]{enhanced,breakable,colback=ReviewPink,colframe=ReviewRed,
 boxrule=0.6pt,arc=1mm,left=2mm,right=2mm,top=1mm,bottom=1mm,
 title={ORIGINAL WORDING -- REVIEW: #1},fonttitle=\small\bfseries,
 before skip=7pt,after skip=8pt}
\newtcolorbox{statusnote}{enhanced,breakable,colback=blue!3!white,colframe=blue!40!black,
 boxrule=0.6pt,arc=1mm,fontupper=\small,left=2mm,right=2mm}
\begin{document}
\raggedbottom

\maketitle

\begin{abstract}
We reproduce the central claims of \citet{inp_former_2025}, which introduces INP-Former, a reconstruction-based method for multi-class industrial anomaly detection. INP-Former extracts Intrinsic Normal Prototypes (INPs) from each test image and uses them to guide a decoder that reconstructs normal features, without a stored memory bank. Using the authors' released code, we reproduce the reported performance on MVTec AD and VisA and extend the evaluation to MVTec AD 2, where image AUROC with the default settings is 63.4, compared with 99.7 on MVTec AD. On MVTec AD, a five-seed ablation shows that most of the performance comes from the INP mechanism, while the two auxiliary losses add small changes whose direction depends on the metric. A paired comparison with both losses disabled confirms the benefit of INP on MVTec AD 2. The default of six prototypes transfers less well: the plateau beyond four prototypes reproduces on MVTec AD and VisA, but on MVTec AD 2 a single prototype outperforms six in AU-PRO on the public split and on both hidden test splits. We also measure practical trade-offs in resolution, backbone size, training length, and LoRA adaptation, and examine performance by defect size, lighting condition, and a stricter AU-PRO limit. Several conclusions from the original benchmarks, including the default prototype count, do not carry over to harder data.

The code we used for reproducing the study can be found here: \url{https://github.com/garasnote/Repro-INPFormer}
\end{abstract}

\section{Introduction}
\label{sec:intro}

Industrial anomaly detection (AD) aims to identify defective products in manufacturing using only normal training samples. Recent reconstruction-based approaches learn to encode and reconstruct normal patterns; at test time, anomalies produce high reconstruction error and are thus detected. A key challenge in multi-class settings where a single model handles all product categories is that the model must capture diverse normal appearances without conflating them.

\citet{inp_former_2025} propose INP-Former, which addresses this by extracting \emph{Intrinsic Normal Prototypes} (INPs) directly from test images via cross-attention aggregation, rather than relying on stored memory banks or class-specific models. The INPs capture the dominant normal patterns present in each test image and guide a prototype-based decoder to reconstruct features. Anomalous regions, which deviate from the extracted normal prototypes, yield high reconstruction error. The method additionally introduces a soft-mining loss $\mathcal{L}_{sm}$, which reweights the gradient contribution of hard-to-reconstruct normal patches, and a coherence loss $\mathcal{L}_c$, which keeps the extracted prototypes close to the encoder features they are drawn from.

In this reproducibility study, we:
\begin{enumerate}
    \item Verify the main performance claims on MVTec AD~\citep{bergmann2019mvtec} and VisA~\citep{zou2022visa} using the authors' released code and pretrained DINOv2 backbone~\citep{oquab2024dinov2}.
    \item Extend the evaluation to MVTec AD 2~\citep{mvtecad2_2024}, a newer and more challenging benchmark.
    \item Extend the loss and module ablation of the original paper with additional configurations and multiple random seeds on MVTec AD.
    \item Conduct additional ablations on the number of INP tokens, input resolution, and backbone size, on MVTec AD, VisA, and MVTec AD 2 as indicated per experiment.
    \item Report analyses beyond the original paper: parameter-efficient adaptation with LoRA~\citep{hu2022lora}, anomaly area versus detection performance, inference speed benchmarks, convergence behaviour, AU-PRO evaluated at a stricter false-positive-rate limit on AD2, and a lighting-condition analysis on AD2.
\end{enumerate}

We organize our experiments around four questions. We first ask whether the original findings reproduce on the benchmarks the authors used (RQ1), and then whether they hold on the harder MVTec AD 2 dataset (RQ2). The remaining analyses ask what it costs to run the model in practice (RQ3) and under what conditions it is fragile (RQ4). So we first reproduce the experiment with the first question,  while the other three test how far the conclusions extend.

\section{Scope and Research Questions}
\label{sec:scope}

We reproduce the original paper's multi-class benchmark results on
MVTec AD and VisA, together with selected module, loss, and
prototype-count analyses. We do not reproduce its Real-IAD,
single-class, few-shot, or zero-shot experiments. Our MVTec AD 2
experiments extend the original study.

\subsection{Claims Under Test}

We investigate the following claims from \citet{inp_former_2025}:

\begin{description}
\item[Claim 1:] INP-Former achieves state-of-the-art multi-class anomaly detection performance, reporting 99.7\% image-level AUROC on MVTec AD and 98.9\% on VisA with a single unified model.
\item[Claim 2:] The INP mechanism is essential: removing prototype aggregation from the decoder substantially degrades performance.
\item[Claim 3:] Both the soft-mining loss $\mathcal{L}_{sm}$ and coherence loss $\mathcal{L}_c$ contribute to the final performance. We assess the changes associated with each term for both detection and localization.
\item[Claim 4:] The number of INP tokens $M$ affects performance. The original paper reports that performance stabilizes beyond four prototypes and adopts $M=6$ as its default on MVTec AD and VisA.
\end{description}

\subsection{Research Questions}

We group the experiments testing the aforementioned claims and other additional analyses under four questions.

\begin{description}
\item[RQ1:] Which of the original INP-Former findings can be reproduced?
\item[RQ2:] Which findings generalize to the more challenging MVTec AD 2 benchmark?
\item[RQ3:] What practical trade-offs affect deployment?
\item[RQ4:] Under what conditions is INP-Former fragile?
\end{description}

\begin{table}[H]
\caption{A table showing which of the research questions address which claims.}
\label{tab:claim_map}
\small
\begin{tabularx}{\linewidth}{@{}llX@{}}
\toprule
Claim & Question & Evidence in this study \\
\midrule
Claim 1 & RQ1 & Main reproduction on MVTec AD and VisA (Section~\ref{sec:results_main}). \\
Claim 2 & RQ1, RQ2 & Loss and module ablation on MVTec AD (Section~\ref{sec:results_main}); paired INP-only versus No-INP comparison on the MVTec AD 2 public split (Section~\ref{sec:ad2_pair}). \\
Claim 3 & RQ1 & Five-configuration loss ablation on MVTec AD (Section~\ref{sec:results_main}). \\
Claim 4 & RQ1, RQ2 & Prototype-count sweeps on MVTec AD and VisA (Section~\ref{sec:results_inp_count}) and on MVTec AD 2 (Section~\ref{sec:rq2}). \\
\bottomrule
\end{tabularx}
\end{table}

We were able to reproduce Claim 1: our MVTec AD scores match the original within 0.1 points, and on VisA the two AUROCs match while our AU-PRO is 0.87 points higher. Removing the INP mechanism lowers performance on both MVTec AD and MVTec AD 2 (both auxiliary losses disabled), which supports Claim 2. Claim 3 holds only in part: the losses change the scores by small amounts whose direction depends on the metric, and the full model tops neither metric in our ablation. We reproduce Claim 4 on MVTec AD and VisA: performance broadly plateaus beyond four prototypes. This pattern does not extend to MVTec AD 2, where a single prototype achieves the highest image AUROC and AU-PRO in our sweep.

\section{Methodology}
\label{sec:methodology}

\subsection{Model Description}

INP-Former uses a frozen DINOv2 encoder with registers~\citep{oquab2024dinov2,darcet2024registers} (ViT-Base/14 by default) to extract multi-layer patch features from input images. The $M$ learnable prototype queries aggregate information from the fused encoder features via cross-attention. The fused features also pass through a bottleneck MLP to form the decoder input. The queries run on each image separately, so every image gets its own $M$ prototypes. The tokens then guide a prototype-guided decoder of eight transformer blocks, in which decoder states act as queries and the aggregated tokens as keys and values. At inference, anomaly maps are computed as the cosine distance between encoder and decoder features, followed by Gaussian smoothing.

We train the model with a cosine-similarity reconstruction objective, using soft mining to reweight its gradients and a coherence term to encourage representative prototypes. The soft-mining loss $\mathcal{L}_{sm}$ reweights the gradient contribution of the reconstruction objective through an exponent $y$, so that hard-to-reconstruct normal patches carry more weight. The coherence loss $\mathcal{L}_c$ is the mean nearest-prototype cosine distance over the fused encoder features, so it encourages each image feature to lie close to at least one prototype.

Our No-INP control bypasses the aggregation step and supplies the full pre-bottleneck encoder-feature sequence as decoder keys and values. Decoder cross-attention is retained. The control is therefore not a self-attention decoder, and we do not claim that it is identical to the ablation reported by \citet{inp_former_2025}. Table~\ref{tab:mechanism_record} records the configurations we vary.

Our reproduction is based on the authors' released implementation, with the experimental controls and evaluation changes described below.

\begin{table}[H]
\caption{Configurations varied in our experiments.}
\label{tab:mechanism_record}
\small
\begin{tabularx}{\linewidth}{@{}p{0.25\linewidth}X@{}}
\toprule
Configuration & Description \\
\midrule
INP aggregation & $M$ learned queries cross-attend to the fused encoder features of each image, before the decoder bottleneck. \\
No-INP control & Bypasses aggregation and supplies the full pre-bottleneck encoder-feature sequence as decoder keys/values. Cross-attention remains; this is not vanilla self-attention, and equivalence to the original authors' control is not established. \\
Soft mining & The exponent $y$ controls gradient reweighting of the cosine reconstruction objective. At $y=0$, reweighting is disabled; reconstruction remains. \\
Coherence term & Mean nearest-prototype cosine distance over fused encoder features. $\lambda=0$ removes its contribution to the training objective. \\
\bottomrule
\end{tabularx}
\end{table}

\subsection{Datasets}

\begin{table}[H]
\caption{Datasets and evaluation scope. The prototype-count sweep uses the public MVTec AD 2 test split; official private scores are reported for $M=1$ and $M=6$.}
\label{tab:dataset_scope}
\small
\begin{tabularx}{\linewidth}{@{}p{0.25\linewidth}X@{}}
\toprule
Dataset & Role in this study \\
\midrule
MVTec AD & 15 categories of industrial products and textures; 3,629 normal training images and 1,725 test images. Original-benchmark reproduction~\citep{bergmann2019mvtec}. \\
VisA & 12 categories of complex structures; 8,659 training images and 2,162 test images. Original-benchmark reproduction~\citep{zou2022visa}. \\
MVTec AD 2 & 8 categories, including Fabric; models trained on the AD2 training split and evaluated on the public test split~\citep{mvtecad2_2024}. The $M=1$ and $M=6$ submissions also have official private scores. Not evaluated in the original paper. \\
Real-IAD & Not included among our reproduced training or evaluation benchmarks~\citep{realiad2024} due to training cost limitations. \\
\bottomrule
\end{tabularx}
\end{table}

\subsection{Experimental Setup and Hyperparameters}
\label{sec:setup}

We follow the hyperparameters of the released code: images are resized to 448 and center-cropped to 392, with batch size 16 and 200 training epochs. We use StableAdamW (learning rate $10^{-3}$, $\beta=(0.9,0.999)$, weight decay $10^{-4}$, AMSGrad, $\epsilon=10^{-10}$) with a warm cosine schedule decaying to $10^{-4}$ after 100 warmup iterations. We clip gradients to a maximum norm of 0.1 at each optimization step. Unless stated otherwise, $M=6$, $y=3$, and $\lambda=0.2$. The DINOv2 encoder is frozen; the bottleneck, learned INP tokens, INP extractor, and decoder are optimized. On MVTec AD 2, each complete test image is resized to 392 and the anomaly map is restored to native size. Experiments used NVIDIA L4 and A100-SXM4-40GB GPUs. The reported FPS measurements used an L4. The resolution and backbone benchmarks used L4 GPUs, including the reported FPS measurements; the paired AD2 runs used A100 GPUs. Table~\ref{tab:paired_settings} lists the settings specific to the paired AD2 comparison.

\begin{table}[H]
\caption{Settings specific to the paired INP-only and No-INP comparison on MVTec AD 2 (Section~\ref{sec:ad2_pair}). Unlisted settings follow Section~\ref{sec:setup}.}
\label{tab:paired_settings}
\small
\begin{tabularx}{\linewidth}{@{}p{0.27\linewidth}X@{}}
\toprule
Setting & Value \\
\midrule
Seed & 1 for both runs. \\
Loss controls & $y=0$ and $\lambda=0$ in both runs; the reconstruction objective remains active. The two training configurations differ only in whether prototype aggregation is used. \\
Prototypes & $M=6$ for INP-only; the No-INP control has no aggregation step. \\
Training hardware & NVIDIA A100-SXM4-40GB for both runs. \\
\bottomrule
\end{tabularx}
\end{table}

\subsection{Evaluation Metrics}

We report image-level AUROC (I-AUROC) for detection, pixel-level AUROC (P-AUROC) for segmentation, and per-region overlap AU-PRO for localization quality. Benchmark metrics are computed per category and averaged without weighting. For all locally scored AD2 public-split experiments, the image-level anomaly score is the mean of the highest-scoring 1\% of pixels in the Gaussian-smoothed $256\times256$ anomaly map. The pooled anomaly-size statistics and the matched-category lighting comparisons are described separately.

For MVTec AD and VisA, we used the evaluation code released with INP-Former to reproduce the original protocol, which integrates AU-PRO to a false-positive rate of 0.30. Our initial AD2 analyses used that evaluation code together with a separate script for low-FPR AU-PRO, and their results were not directly comparable across tables. We therefore re-evaluated the AD2 public split with the official MVTec AU-PRO implementation recommended for AD2's local evaluation, using the same prediction maps at false-positive-rate limits of 0.05 and 0.30. Public AD2 benchmark scores use the unweighted mean across all eight categories. Official private and private-mixed scores are obtained separately from the benchmark server. Lighting comparisons use condition-matched images and the same categories on both sides of each comparison.

\section{RQ1: Which original INP-Former findings reproduce?}
\label{sec:rq1}

\label{sec:results}

\subsection{Main MVTec AD and VisA Reproduction}
\label{sec:results_main}

Table~\ref{tab:main_results} compares our reproduced results with those reported in the original paper. Per-category results are provided in Appendix~\ref{app:per_category}.

\begin{table}[H]
\caption{Multi-class anomaly detection on MVTec AD and VisA. ``Original'' indicates the results reported by \citet{inp_former_2025}; ``Ours'' gives the reproduced results from our retraining of their models. This run is separate from the five-seed ablation in Table~\ref{tab:ablation_loss}. All entries are percentages.}
\label{tab:main_results}
\begin{center}
\begin{tabular}{llccc}
\toprule
Dataset & Source & I-AUROC & P-AUROC & P-AUPRO \\
\midrule
\multirow{2}{*}{MVTec AD} & Original & 99.7 & 98.5 & 94.9 \\
                           & Ours     & 99.67 & 98.45 & 94.95 \\
\midrule
\multirow{2}{*}{VisA}     & Original & 98.9 & 98.9 & 94.4 \\
                           & Ours     & 98.90 & 98.97 & 95.27 \\
\bottomrule
\end{tabular}
\end{center}
\end{table}

Our MVTec AD results match the original to within 0.1 percentage points across all three metrics, and VisA image and pixel AUROCs differ by 0.00 and 0.07 points, respectively; VisA AU-PRO is 0.87 points higher than the published rounded value. This supports reproduction of the reported benchmark performance. MVTec AD 2 was not evaluated in the original paper, so it has no ``Original'' row; we report it separately in Section~\ref{sec:rq2}.

% Yunfei: doesn't it make more sense to have all the values in one table? Or is separating them for visually clear comparison that important?

Table~\ref{tab:ablation_loss} compares our loss and module ablations on MVTec AD with those reported in the original publication.

\begin{table}[H]
\caption{Loss and module ablation on MVTec AD in percentage. ``Original'' values are from Table~5 of \citet{inp_former_2025}; ``Ours'' are the means and standard deviations over five seeds. No-INP, INP+$\mathcal{L}_{sm}$, and Full come from an earlier set of runs than INP-only and INP+$\mathcal{L}_c$, which use seeds 1, 2, 3, 42, and 123.}
\label{tab:ablation_loss}
\begin{center}
\small
\resizebox{\linewidth}{!}{%
\begin{tabular}{lcccccccc}
\toprule
 & & & & \multicolumn{2}{c}{Original} & \multicolumn{2}{c}{Ours} \\
\cmidrule(lr){5-6} \cmidrule(lr){7-8}
Configuration & INP & $\mathcal{L}_c$ & $\mathcal{L}_{sm}$ & I-AUROC & P-AUPRO & I-AUROC & P-AUPRO \\
\midrule
No INP & \texttimes & \texttimes & \texttimes & 98.59 & 92.73 & $98.19 \pm 0.65$ & $91.55 \pm 1.19$ \\
INP only & \checkmark & \texttimes & \texttimes & 99.53 & 94.69 & $99.60 \pm 0.03$ & $95.14 \pm 0.05$ \\
INP + $\mathcal{L}_c$ & \checkmark & \checkmark & \texttimes & 99.61 & 95.10 & $99.64 \pm 0.02$ & $95.23 \pm 0.10$ \\
INP + $\mathcal{L}_{sm}$$^\dagger$ & \checkmark & \texttimes & \checkmark & --- & --- & $99.64 \pm 0.04$ & $94.85 \pm 0.08$ \\
Full           & \checkmark & \checkmark & \checkmark & 99.67 & 94.87 & $99.62 \pm 0.01$ & $94.92 \pm 0.07$ \\
\bottomrule
\end{tabular}}%
\end{center}
\vspace{-0.5em}
{\footnotesize $^\dagger$ Configuration not present in the original paper's ablation.}
\end{table}

When we disable both auxiliary losses, the reported INP-only configuration exceeds the No-INP control by 1.41 points in image AUROC and 3.59 points in AU-PRO. The INP-enabled configurations also show less variation across seeds in the reported results. These observations are consistent with a benefit from prototype aggregation, and they support Claim 2. Because the No-INP row comes from an earlier set of runs than the new configurations, we examine the contribution of aggregation further in the paired AD2 experiment of Section~\ref{sec:ad2_pair}, where the two training configurations differ only in whether the decoder receives the extracted prototypes or the full image-feature sequence.

Our full configuration closely reproduces the original paper's MVTec AD results. The original ablation evaluated No-INP, INP-only, INP with coherence loss, and the full model. We add a soft-mining-only configuration, which the original ablation did not include.

The additional runs show that INP-only accounts for most of the final performance. The coherence loss raises mean image AUROC by 0.04 points and AU-PRO by 0.09 points. Among our reported configurations, INP with coherence loss has the highest mean AU-PRO, exceeding the full model by 0.31 points. The soft-mining-only configuration also reaches a high image AUROC, but its AU-PRO is lower than that of INP-only.

The two auxiliary losses therefore do not produce consistent improvement across the reported metrics on MVTec AD. The full configuration performs well, but does not obtain the highest mean on either metric in our table. However, we don't interpret this to mean that detection and localization are fully separate: the original paper's VisA results show that adding soft mining to the coherence-loss configuration improves both image AUROC and AU-PRO. The usefulness of each loss has to be assessed for the dataset and metric of interest.

\subsection{Prototype Count on MVTec AD and VisA}
\label{sec:results_inp_count}

% Yunfei: wouldn't it be better to have the original paper's results also here, and for us to have a table that shows the contrast?
\begin{table}[H]
\centering\small
\caption{Prototype-count sweep on MVTec AD and VisA (percent). Results are from one seed. AU-PRO is integrated up to a false-positive rate of 0.30.}
\label{tab:prototype_ad_visa}
\begin{tabular}{rccc ccc}\toprule
 & \multicolumn{3}{c}{MVTec AD} & \multicolumn{3}{c}{VisA}\\
$M$ & I-AUROC & P-AUROC & AU-PRO & I-AUROC & P-AUROC & AU-PRO\\\midrule

1 & 98.37 & 96.54 & 92.21 & 97.34 & 97.66 & 93.00 \\

2 & 99.36 & 98.12 & 94.39 & 98.15 & 98.53 & 93.80 \\

4 & 99.55 & 98.39 & 94.75 & 98.81 & 98.84 & 95.04 \\

6 & 99.67 & 98.45 & 94.95 & 98.90 & 98.97 & 95.27 \\

8 & 99.60 & 98.42 & 94.90 & 98.84 & 98.98 & 95.23 \\

12 & 99.67 & 98.51 & 95.12 & 98.67 & 98.94 & 95.07 \\

16 & 99.63 & 98.53 & 95.08 & 98.72 & 98.88 & 94.85 \\

\bottomrule\end{tabular}
\end{table}

The original paper reports that performance stabilizes beyond four prototypes and selects six as its default. Our MVTec AD and VisA sweeps broadly reproduce this pattern (Table~\ref{tab:prototype_ad_visa}): most of the improvement occurs between one and four prototypes, with comparatively small changes thereafter. No single prototype count is optimal across both datasets and all metrics, neither does increasing the count lead to a consistent decline. Our findings indicate a general plateau rather than a unique optimum at six. Our experiments with MVTec AD 2 dataset sweep produces a different pattern reported in Section~\ref{sec:rq2}.

\section{RQ2: Which findings generalize to MVTec AD 2?}
\label{sec:rq2}

\subsection{Motivation and Corrected Public Evaluation}

We additionally report results on MVTec AD 2, which was created in response to the saturation of established benchmarks such as MVTec AD and VisA. The dataset includes industrial conditions that were underrepresented in earlier benchmarks, including lighting-induced distribution shifts and defects at image borders, with each of its eight categories captured under at least four lighting conditions. We include MVTec AD 2 to test whether our findings persist after retraining INP-Former under these more challenging conditions. Our corrected evaluation covers all eight categories of the public test split and therefore represents neither zero-shot transfer nor evaluation on the hidden-split private benchmark. Table~\ref{tab:inp_count} also reports the official hidden-split scores.

% Yunfei shouldn't table 9 just be merged with the MVTEC AD1 and Visa results as additional columns?

\begin{table}[H]\centering\small
\caption{Prototype-count sweep on MVTec AD 2 (percent), full model, seed 1. Public-split scores are computed locally. Private and private-mixed scores come from the official benchmark server, to which we submitted $M=1$ and $M=6$ only due to submission limits.}
\label{tab:inp_count}
\begin{tabular}{rcc ccc c}\toprule
 & & & \multicolumn{3}{c}{AU-PRO@0.05} & \\
\cmidrule(lr){4-6}
$M$ & I-AUROC & P-AUROC & Public & Private & Priv.-mixed & AU-PRO@0.30\\\midrule
1  & 65.75 & 87.68 & 32.12 & 37.14 & 33.98 & 57.39 \\
2  & 64.68 & 87.45 & 27.68 & ---   & ---   & 55.09 \\
4  & 63.96 & 87.58 & 25.06 & ---   & ---   & 53.75 \\
6  & 63.39 & 87.98 & 26.67 & 34.51 & 28.46 & 54.86 \\
8  & 62.48 & 87.21 & 22.94 & ---   & ---   & 52.47 \\
12 & 63.71 & 86.99 & 22.56 & ---   & ---   & 52.01 \\
16 & 62.56 & 86.95 & 22.87 & ---   & ---   & 52.58 \\
\bottomrule\end{tabular}\end{table}

On MVTec AD 2, image AUROC ranges from 62.48 to 65.75 across the sweep, compared with 99.67 on MVTec AD (Table~\ref{tab:main_results}).

The official private evaluations also favor $M=1$ over $M=6$ in AU-PRO@0.05 in the hidden splits separate from the public dataset. Appendix~\ref{app:ad2_full_categories} reports the full-model public category scores. The $M=1$ model has higher AU-PRO@0.05 in seven of eight categories, whereas image AUROC favors each model in four categories.

Table~\ref{tab:ad2_baseline_context} places our hidden-split
results alongside published MVTec AD 2 evaluations of
EfficientAD-S~\citep{batzner2024efficientad} and
PatchCore~\citep{roth2022towards}. The baseline scores
come from \citet{mvtecad2_2024} at $256\times256$ and
$512\times512$; our test images are resized to $392\times392$.
Differences in training settings and input size limit direct
comparisons between methods.

\subsection{Prototype-Count Behavior Across Datasets}

On MVTec AD and VisA, performance improves until four to six prototypes and then remains relatively stable, consistent with the plateau reported in the original paper. In contrast, the MVTec AD 2 public sweep achieves its highest I-AUROC and localization scores with a single prototype, while P-AUROC varies within a narrower range and is highest at $M=6$. The M value is allocated to each image and is category independent, so the number of categories in each of the datasets does not explain this difference. Although the mechanism behind this difference remains unsolved right now, the practical implication is clear: the prototype count should be validated on the target benchmark rather than treated as a universal default.

\subsection{Architecture Comparison with Both Auxiliary Losses Disabled}
\label{sec:ad2_pair}

Table~\ref{tab:ad2_architecture_pair} compares two runs that differ only in whether the decoder receives aggregated prototypes or the full encoder-feature sequence. We disabled the auxiliary losses in both runs to compare the impact of the INP mechanism directly without the loss terms. Per-category differences are given in Appendix~\ref{app:ad2_pair_categories}.

% Yunfei: for AD2, why am I describing all of these instead of describing them in a table or with the other settings or something? It seems like the changes are justified in magnitude... the part just seems weird in general
\begin{table}[H]
\centering
\caption{Paired INP-only and No-INP results on the eight-category MVTec AD 2 public test split (percent), seed 1. Both runs use 200 training epochs with the auxiliary losses disabled; reconstruction remains active. Training uses 448/392 resize/crop. Whole test images are resized to 392, with prediction maps restored to native size. INP-only uses $M=6$; No-INP feeds the full feature sequence to the decoder. The control retains decoder cross-attention, and its equivalence to the original authors' control is not established. Deltas are calculated before rounding.}
\label{tab:ad2_architecture_pair}
\small
\begin{tabular}{lrrrr}
\toprule
Configuration & I-AUROC & P-AUROC & AU-PRO@0.05 & AU-PRO@0.30 \\
\midrule
No-INP control & 59.90 & 86.08 & 18.98 & 48.20 \\
INP-only & 64.79 & 88.04 & 30.03 & 56.48 \\
$\Delta$ (INP minus control) & +4.89 & +1.96 & +11.05 & +8.29 \\
\bottomrule
\end{tabular}
\end{table}

When we disable both auxiliary losses, the INP-only model exceeds the No-INP control by 4.89 points in image AUROC and 11.05 points in AU-PRO@0.05 on the AD2 public split. The control allows the decoder to process the full feature sequence, compared to the INP-only model, which aggregates that information into image-specific prototypes. Our comparison supports the usefulness of prototype aggregation under the tested configuration. Exceptions do exist, however. The AU-PRO@0.05 of the INP model drops by 6.02 points for can, and image AUROC falls by 7.83 points for fruit jelly.

\section{RQ3: What practical trade-offs affect deployment?}
\label{sec:rq3}

\label{sec:extensions}

\subsection{Resolution and Backbone Ablation}

Table~\ref{tab:resolution} shows the effect of input resolution and backbone size on MVTec AD. Higher resolution provides consistent improvements, with P-AUPRO benefiting most (+2.9 points from 224 to 448). ViT-Small/14 loses 0.47 I-AUROC and 0.56 P-AUPRO relative to ViT-Base/14 at the same resolution. The corresponding throughput figures are given in Section~\ref{sec:inference}.

\begin{table}[H]
\caption{Resolution and backbone ablation on MVTec AD (percent). The ViT-Base/14 448/392 row is shown in both comparisons; it is the same run.}
\label{tab:resolution}
\begin{center}
\begin{tabular}{llccc}
\toprule
Backbone & Input/Crop & I-AUROC & P-AUROC & P-AUPRO \\
\midrule
ViT-Base/14 & 224/196 & 99.21 & 97.97 & 92.07 \\
ViT-Base/14 & 336/294 & 99.63 & 98.28 & 94.04 \\
ViT-Base/14 & 448/392 & 99.67 & 98.45 & 94.95 \\
\midrule
ViT-Small/14 & 448/392 & 99.20 & 98.18 & 94.39 \\
ViT-Base/14  & 448/392 & 99.67 & 98.45 & 94.95 \\
\bottomrule
\end{tabular}
\end{center}
\end{table}

On MVTec AD, these results favor a smaller backbone over a lower input resolution when faster inference is needed. ViT-Small at 448/392 runs at 72.6 FPS and loses 0.56 AU-PRO points; ViT-Base at 224/196 runs at 85.3 FPS and loses 2.88.

\subsection{Inference Speed}
\label{sec:inference}

We measure inference speed on a single NVIDIA L4 GPU (Table~\ref{tab:inference}). Resolution matters most: going from 224 to 448 makes the model 3.4 times slower. ViT-Small is 2.9 times faster than ViT-Base at the same resolution. The number of prototypes has almost no effect, with speed staying between 25.0 and 25.2 FPS for every $M$ from 2 to 16.

\begin{table}[H]
\caption{Model-forward inference speed on an NVIDIA L4 GPU at batch size 1 (50 warmup and 200 timed iterations). Timing excludes image preprocessing and output-map postprocessing.}
\label{tab:inference}
\begin{center}
\begin{tabular}{llcc}
\toprule
Config & Setting & FPS & Latency (ms) \\
\midrule
\multirow{3}{*}{Resolution (ViT-B, $M$=6)} & 224/196 & 85.3 & 11.7 \\
 & 336/294 & 50.1 & 20.0 \\
 & 448/392 & 25.0 & 40.0 \\
\midrule
\multirow{2}{*}{Backbone (448/392, $M$=6)} & ViT-Small/14 & 72.6 & 13.8 \\
 & ViT-Base/14 & 25.0 & 40.0 \\
\midrule
INP count (ViT-B, 448/392) & $M$=2--16 & 25.0--25.2 & 39.7--40.0 \\
\bottomrule
\end{tabular}
\end{center}
\end{table}

\subsection{Parameter-Efficient Adaptation}
\label{sec:lora}

We test whether INP-Former can be adapted to a new dataset by training only a small part of the model. We use LoRA~\citep{hu2022lora}, which attaches small trainable matrices to existing layers so that the original weights stay fixed. Starting from an INP-Former checkpoint trained on Real-IAD~\citep{realiad2024} released by the authors, we add adapters of rank $r \in \{2, 4, 8\}$ to the linear layers of the bottleneck and decoder (42 adapters in total). We train these adapters together with the prototype tokens and aggregation module for 50 epochs on MVTec AD and on VisA, and keep all other parameters frozen. Table~\ref{tab:lora} lists the trainable share of the model for each rank. Most of it is the prototype tokens and aggregation module; the adapters alone account for 0.14\% to 0.58\%.

\begin{table}[H]
\caption{Parameter-efficient adaptation on MVTec AD and VisA (percent). ``Zero-shot'' is the Real-IAD checkpoint evaluated on the target dataset without adaptation; ``Full train'' is the 200-epoch model from Table~\ref{tab:main_results}. ``Total'' counts the adapters, prototype tokens, and aggregation module. The encoder is frozen in all rows.}
\label{tab:lora}
\begin{center}
\small
\resizebox{\linewidth}{!}{%
\begin{tabular}{lcccccccc}
\toprule
 & \multicolumn{2}{c}{Trainable (\%)} & \multicolumn{3}{c}{MVTec AD} & \multicolumn{3}{c}{VisA} \\
\cmidrule(lr){2-3} \cmidrule(lr){4-6} \cmidrule(lr){7-9}
Method & Total & Adapters & I-AUROC & P-AUROC & P-AUPRO & I-AUROC & P-AUROC & P-AUPRO \\
\midrule
Zero-shot & --- & --- & 69.15 & 79.92 & 57.17 & 51.47 & 72.60 & 31.60 \\
LoRA $r=2$ & 4.71 & 0.14 & 97.22 & 96.21 & 91.30 & 91.79 & 96.29 & 85.11 \\
LoRA $r=4$ & 4.85 & 0.29 & 97.81 & 96.62 & 91.87 & 93.17 & 96.79 & 87.55 \\
LoRA $r=8$ & 5.12 & 0.58 & 98.16 & 96.96 & 92.47 & 94.44 & 97.25 & 89.91 \\
\midrule
Full train & --- & --- & 99.67 & 98.45 & 94.95 & 98.90 & 98.97 & 95.27 \\
\bottomrule
\end{tabular}}%
\end{center}
\end{table}

Adaptation recovers most of the full-training performance on both datasets. At rank 8, with 5.12\% of the parameters trainable and a quarter of the training epochs, image AUROC is 1.51 points below full training on MVTec AD and 4.46 points below on VisA. Each increase in rank monotonically improves every metric in the table.

\subsection{Convergence Behavior}
\label{sec:epoch_efficiency}

We evaluate how quickly INP-Former converges by training the full configuration ($y=3$, $\lambda=0.2$) on MVTec AD and evaluating at intermediate checkpoints of a single run (Table~\ref{tab:epoch_eff}).

\begin{table}[H]
\caption{Training convergence on MVTec AD. All metrics are means across 15 categories. $\Delta$ shows the difference in percentage points from epoch 200 (final), separate from the results in Table~\ref{tab:main_results}.}
\label{tab:epoch_eff}
\begin{center}
\begin{tabular}{rccccr}
\toprule
Epoch & Loss & I-AUROC & P-AUROC & P-AUPRO & $\Delta$P-AUPRO \\
\midrule
25  & 0.163 & 99.29 & 98.07 & 94.57 & $-0.39$ \\
50  & 0.145 & 99.50 & 98.26 & 94.76 & $-0.20$ \\
100 & 0.133 & 99.55 & 98.40 & 94.90 & $-0.06$ \\
150 & 0.128 & 99.59 & 98.45 & 94.95 & $-0.01$ \\
200 & 0.126 & 99.61 & 98.48 & 94.96 & --- \\
\bottomrule
\end{tabular}
\end{center}
\end{table}

Most of the improvement happens early: at epoch 25, all three metrics are already within half a point of their final values. After epoch 100, no metric changes by more than 0.1 points, even though the training loss keeps falling. These scores come from the test set of a single run to illustrate the trend that most of the learning may happen in the earlier epochs.

\Needspace{14\baselineskip}
\section{RQ4: Under what conditions is INP-Former fragile?}
\label{sec:rq4}

\subsection{Anomaly Size on MVTec AD and VisA}
\label{sec:area}

INP-Former assumes that most of a defective image is still normal, so the prototypes extracted from it mostly describe normal content. We therefore expected large anomalies to be harder to detect, since they leave fewer normal patches for the prototypes to draw on. To test this, we compare each anomalous image's score with the area of its ground-truth defect, pooling all categories within each dataset. Area is measured on the ground-truth mask after resizing to $256\times256$.

Anomaly scores increase with defect area: Spearman $\rho \approx +0.53$ on MVTec AD (1,258 anomalous images) and $\rho \approx +0.79$ on VisA (1,200 anomalous images). For MVTec AD, we also count how many anomalous images score above a threshold set at the mean plus two standard deviations of the normal test images' scores. Among the 250 images with the smallest defects, 63.6\% do.

These results do not support our original hypothesis. Since scores rise with area, small defects are the ones most likely to be missed; about a third of the smallest MVTec AD defects fall below the threshold. The correlations pool all categories, however, and categories differ in both typical defect size and difficulty, so part of the trend may reflect category rather than size.

\subsection{Lighting Conditions on MVTec AD 2}
\label{sec:lighting}

MVTec AD 2 photographs each object under several lighting conditions, and the overall scores mix them together. To see how each condition affects the model, we split the existing public-split predictions of the $M=1$ and $M=6$ full models by condition and compare each condition with regular lighting. Each comparison uses the same categories on both sides, with normal and anomalous images drawn from the same condition. Fruit jelly has no \texttt{shift\_2} or \texttt{shift\_3} images, and only vial has \texttt{shift\_4}.

\begin{table}[H]
\centering\small
\caption{Change from regular lighting on the AD2 public split (percentage points), full models, seed 1. Negative values mean lower scores under the listed condition. Scores are computed within each condition.}
\label{tab:lighting_matched}
\begin{tabular}{lrrrrr}
\toprule
 & & \multicolumn{2}{c}{I-AUROC change} & \multicolumn{2}{c}{AU-PRO@0.05 change}\\
Condition & Categories evaluated & $M=1$ & $M=6$ & $M=1$ & $M=6$\\
\midrule
overexposed & 8 & +0.73 & +1.11 & -0.28 & +0.49 \\
underexposed & 8 & -2.61 & -4.04 & -3.05 & -4.70 \\
shift\_1 & 8 & -5.07 & -1.72 & -4.28 & -9.26 \\
shift\_2 & 7 & -0.24 & -0.15 & +0.46 & -3.40 \\
shift\_3 & 7 & -6.72 & -2.53 & -1.43 & -5.90 \\
shift\_4 & 1 & +6.67 & +12.00 & -8.88 & -13.15 \\
\bottomrule
\end{tabular}
\end{table}

Each category contributes 15 anomalous and 4 to 12 normal images per condition. The same scenes recur across conditions, and the same condition name can mean a different lighting change in different categories. Appendix~\ref{app:lighting_full} gives the full per-category results.

Splitting by condition shows that the two models respond differently. $M=6$ holds image AUROC better than $M=1$ under most conditions, while $M=1$ holds AU-PRO@0.05 better: under \texttt{shift\_1}, for example, $M=6$ loses 1.72 points of image AUROC against 5.07 for $M=1$, but 9.26 points of AU-PRO@0.05 against 4.28. Underexposure lowers both metrics for both models, while overexposure changes little. With one seed and 4 to 12 normal images per cell, individual differences are noisy but patterns across conditions are more reliable.

\subsection{Localization Under Strict Low-FPR Evaluation}
\label{sec:aupro_strict}

The standard AU-PRO integrates the per-region overlap curve up to a pixel false-positive rate of 0.30. \citet{mvtecad2_2024} argue that this limit tolerates too many false positives for industrial inspection and use 0.05 instead. We report both limits on the MVTec AD 2 public split, computed from the same prediction maps with the official implementation (Tables~\ref{tab:inp_count} and~\ref{tab:ad2_architecture_pair}).

The stricter limit makes the differences between configurations larger without changing which one is better. Across the prototype sweep, the gap between the best and worst setting grows from 5.38 points at 0.30 to 9.56 points at 0.05, and the ranking stays the same except that $M=8$ and $M=16$ swap places. The gap between INP-only and the No-INP control grows in the same way, from 8.29 to 11.05 points. Both limits count false-positive pixels, so they measure how precisely the model localizes defects.

\section{Discussion}
\label{sec:discussion}

Our results reproduce INP-Former's reported performance on MVTec AD and VisA and show that the INP mechanism does most of the work. On MVTec AD, INP-only performs about as well as the full model, and on MVTec AD 2 the paired comparison favors INP by 11.05 points of AU-PRO@0.05. The two auxiliary losses add small gains whose direction depends on the metric. The prototype count transfers less well: the plateau beyond four prototypes holds on MVTec AD and VisA, but on MVTec AD 2 one prototype beats six in AU-PRO@0.05 on the public split and on both hidden splits. Averages also hide differences in stability: $M=1$ scores higher overall, while $M=6$ holds image AUROC better under most lighting changes.

The practical analyses point to cheaper settings on MVTec AD. At similar speed, a smaller backbone at full resolution localizes defects better than a larger backbone at low resolution. Training gains mostly settle by epoch 100, and LoRA adaptation recovers most of the full-training performance with about 5\% of the model trainable. Small defects remain the hardest case.

Our study has several caveats. The MVTec AD 2 experiments use a single seed, and The MVTec AD loss-ablation rows come from separate batches of runs, so we place more weight on the controlled, paired AD2 comparison when assessing the contribution of INPs. Per-run code snapshots are unavailable for these rows. The loss ablation, resolution and backbone study, and convergence analysis cover MVTec AD only. Our comparison with other methods uses published AD2 results; a comparison under matched training and preprocessing settings remains outside the scope of this study.

\section{Conclusion}
\label{sec:conclusion}

We reproduce INP-Former's benchmark performance on MVTec AD and VisA, and find that prototype aggregation accounts for most of it while the auxiliary losses add little on MVTec AD. Performance drops sharply on MVTec AD 2, where image AUROC with the default settings is 63.4 compared with 99.7 on MVTec AD. There, aggregation still helps, but the default of six prototypes does not carry over: one prototype outperforms six in AU-PRO@0.05 on the public split and on both hidden splits. On MVTec AD, a smaller backbone and LoRA adaptation each lower cost at a small loss of accuracy. The model's weak points are small defects and some lighting changes, where the better prototype setting depends on the metric, so findings from the original benchmarks need rechecking on harder data before they guide deployment.

\subsubsection*{Author Contributions}

Both authors contributed roughly equally in both writing and coding.

% \subsubsection*{Acknowledgments}

% Computational resources provided by the FRIDA HPC cluster at the University of Ljubljana.

\bibliography{main}

\bibliographystyle{tmlr}

\appendix

\section{Per-Category Results}
\label{app:per_category}

\begin{table}[H]
\caption{Per-category results on MVTec AD (ViT-Base/14, 448/392, $M$=6).}
\label{tab:mvtec_per_cat}
\begin{center}
\small
\begin{tabular}{lccc}
\toprule
Category & I-AUROC & P-AUROC & P-AUPRO \\
\midrule
carpet     & 99.88 & 99.35 & 97.89 \\
grid       & 99.92 & 99.42 & 97.75 \\
leather    & 100.0 & 99.42 & 98.13 \\
tile       & 100.0 & 97.68 & 88.36 \\
wood       & 99.91 & 97.54 & 93.84 \\
bottle     & 100.0 & 99.06 & 96.90 \\
cable      & 100.0 & 98.80 & 94.78 \\
capsule    & 99.00 & 98.74 & 97.02 \\
hazelnut   & 100.0 & 99.41 & 96.69 \\
metal\_nut & 100.0 & 97.35 & 95.24 \\
pill       & 99.10 & 97.73 & 97.37 \\
screw      & 97.46 & 99.58 & 98.00 \\
toothbrush & 100.0 & 98.98 & 95.75 \\
transistor & 99.71 & 94.73 & 79.65 \\
zipper     & 100.0 & 98.98 & 96.86 \\
\midrule
\textbf{Mean} & \textbf{99.67} & \textbf{98.45} & \textbf{94.95} \\
\bottomrule
\end{tabular}
\end{center}
\end{table}

\section{Paired Public AD2 Category Differences}
\label{app:ad2_pair_categories}
\begin{table}[H]
\centering
\caption{Per-category difference (INP-only minus No-INP) on the MVTec AD 2 public split, in percentage points. Both runs are shown in Table~\ref{tab:ad2_architecture_pair}; negative values favor No-INP.}
\small
\begin{tabular}{lrrrr}
\toprule
Category & I-AUROC & P-AUROC & AU-PRO@0.05 & AU-PRO@0.30 \\
\midrule
can & +2.35 & +1.32 & -6.02 & -0.22 \\
fabric & +9.60 & -2.85 & +10.83 & +17.92 \\
fruit\_jelly & -7.83 & +4.09 & +13.54 & +11.29 \\
rice & +3.62 & -0.56 & +13.92 & -1.77 \\
sheet\_metal & +16.34 & +3.90 & +6.23 & +6.56 \\
vial & +3.37 & -0.11 & +27.42 & +14.48 \\
wallplugs & +11.04 & +8.56 & +6.73 & +12.77 \\
walnuts & +0.67 & +1.31 & +15.74 & +5.26 \\
\bottomrule
\end{tabular}
\end{table}

\section{Full-Model AD2 Public Category Results}
\label{app:ad2_full_categories}
\begin{table}[H]\centering\small
\caption{Full-model AD2 public scores (percent), seed 1, with both auxiliary losses enabled. Each cell is $M=1$ / $M=6$. These category scores underlie Table~\ref{tab:inp_count}; they are distinct from the loss-disabled architecture comparison.}
\setlength{\tabcolsep}{5pt}
\begin{tabular}{lrrrr}\toprule
Category & I-AUROC & P-AUROC & AU-PRO@0.05 & AU-PRO@0.30\\\midrule
can & 50.65 / 52.47 & 64.97 / 69.36 & 2.33 / 1.17 & 22.14 / 24.82 \\
fabric & 56.65 / 69.16 & 83.47 / 90.74 & 21.91 / 22.87 & 43.88 / 51.98 \\
fruit\_jelly & 74.58 / 61.00 & 94.87 / 95.57 & 39.40 / 30.44 & 75.45 / 74.84 \\
rice & 70.42 / 68.76 & 93.93 / 93.51 & 45.85 / 37.11 & 67.05 / 60.25 \\
sheet\_metal & 62.50 / 54.77 & 89.04 / 89.11 & 13.25 / 7.74 & 30.98 / 22.82 \\
vial & 91.65 / 77.39 & 90.14 / 85.76 & 65.40 / 54.72 & 89.39 / 83.30 \\
wallplugs & 44.07 / 46.06 & 88.34 / 83.84 & 19.54 / 14.76 & 48.74 / 40.48 \\
walnuts & 75.50 / 77.52 & 96.70 / 95.95 & 49.27 / 44.58 & 81.52 / 80.37 \\
\midrule
Mean & 65.75 / 63.39 & 87.68 / 87.98 & 32.12 / 26.67 & 57.39 / 54.86 \\
\bottomrule\end{tabular}\end{table}

\subsection{Published AD2 Baseline Context}

\begin{table}[htbp]
\centering
\small
\caption{Hidden-split AU-PRO@0.05 (percent). Baseline scores
are reported by \citet{mvtecad2_2024} for EfficientAD-S
\citep{batzner2024efficientad} and PatchCore
\citep{roth2022towards}. INP-Former scores are our official
server results. Methods use different training settings
and input sizes.}
\label{tab:ad2_baseline_context}
\begin{tabular}{lrrr}
\toprule
Method & Input size & Private & Private-mixed \\
\midrule
EfficientAD-S & $256\times256$ & 30.8 & 19.2 \\
PatchCore & $256\times256$ & 28.8 & 26.0 \\
EfficientAD-S & $512\times512$ & 40.0 & 25.3 \\
PatchCore & $512\times512$ & 41.9 & 35.8 \\
\midrule
INP-Former, $M=1$ & $392\times392$ & 37.14 & 33.98 \\
INP-Former, $M=6$ & $392\times392$ & 34.51 & 28.46 \\
\bottomrule
\end{tabular}
\end{table}

\section{Condition-Level AD2 Lighting Results}
\label{app:lighting_full}
\begin{table}[H]\centering\small
\caption{Matched regular / changed-lighting means (percent), with $M=1$ and $M=6$ adjacent. Each metric cell is regular / changed. Conditions use the same matched categories as Table~\ref{tab:lighting_matched}; its deltas were computed before rounding.}
\setlength{\tabcolsep}{4pt}
\begin{tabular}{lr rrrr}\toprule
Condition & $M$ & I-AUROC & P-AUROC & AU-PRO@0.05 & AU-PRO@0.30\\\midrule
overexposed & 1 & 69.66 / 70.39 & 88.39 / 88.08 & 34.43 / 34.15 & 58.24 / 58.21 \\
 & 6 & 68.21 / 69.33 & 89.98 / 89.60 & 32.75 / 33.25 & 58.16 / 58.01 \\
\addlinespace[2pt]
underexposed & 1 & 69.66 / 67.05 & 88.39 / 87.14 & 34.43 / 31.38 & 58.24 / 56.40 \\
 & 6 & 68.21 / 64.18 & 89.98 / 87.42 & 32.75 / 28.06 & 58.16 / 54.46 \\
\addlinespace[2pt]
shift\_1 & 1 & 69.66 / 64.59 & 88.39 / 88.19 & 34.43 / 30.15 & 58.24 / 57.38 \\
 & 6 & 68.21 / 66.49 & 89.98 / 89.07 & 32.75 / 23.50 & 58.16 / 55.89 \\
\addlinespace[2pt]
shift\_2 & 1 & 67.23 / 66.99 & 87.38 / 85.93 & 33.16 / 33.62 & 55.52 / 56.16 \\
 & 6 & 66.15 / 66.00 & 89.09 / 86.57 & 32.32 / 28.92 & 55.40 / 53.18 \\
\addlinespace[2pt]
shift\_3 & 1 & 67.23 / 60.52 & 87.38 / 87.26 & 33.16 / 31.73 & 55.52 / 57.52 \\
 & 6 & 66.15 / 63.62 & 89.09 / 88.31 & 32.32 / 26.42 & 55.40 / 55.36 \\
\addlinespace[2pt]
shift\_4 & 1 & 88.00 / 94.67 & 90.25 / 92.87 & 70.77 / 61.89 & 90.80 / 89.40 \\
 & 6 & 82.67 / 94.67 & 88.27 / 90.84 & 65.95 / 52.80 & 89.35 / 85.49 \\
\bottomrule\end{tabular}\end{table}

\begin{table}[H]\centering\footnotesize
\caption{Condition-level AD2 public scores (percent). Each metric cell is $M=1$ / $M=6$; $n$ is the normal + anomalous image count shared by both models. Normal images are condition-matched. Conditions absent from a category are unavailable, not zero.}
\label{tab:lighting_all_conditions}
\setlength{\tabcolsep}{3.5pt}
\begin{tabular}{llr rrrr}\toprule
Category & Condition & $n$ & I-AUROC & P-AUROC & PRO@0.05 & PRO@0.30\\\midrule
can & regular & 12+15 & 51.11 / 56.11 & 65.52 / 72.00 & 0.80 / 1.45 & 20.68 / 28.21 \\
 & overexposed & 12+15 & 47.22 / 54.44 & 66.34 / 71.65 & 1.54 / 1.69 & 21.98 / 28.48 \\
 & shift\_1 & 12+15 & 56.67 / 69.44 & 66.90 / 70.67 & 5.60 / 2.11 & 27.29 / 28.54 \\
 & shift\_2 & 12+15 & 51.11 / 45.56 & 62.56 / 66.65 & 0.80 / 0.31 & 18.41 / 18.57 \\
 & shift\_3 & 12+15 & 55.56 / 53.89 & 66.48 / 69.36 & 4.33 / 0.77 & 26.54 / 23.90 \\
 & underexposed & 12+15 & 42.22 / 66.11 & 62.32 / 68.78 & 0.13 / 0.86 & 17.76 / 22.79 \\
\addlinespace[2pt]
fabric & regular & 11+15 & 53.33 / 73.94 & 85.30 / 91.40 & 17.64 / 21.73 & 39.40 / 49.68 \\
 & overexposed & 11+15 & 53.33 / 71.52 & 84.89 / 91.24 & 17.35 / 21.81 & 39.94 / 50.00 \\
 & shift\_1 & 11+15 & 55.15 / 74.55 & 82.97 / 90.05 & 18.07 / 23.07 & 40.11 / 51.84 \\
 & shift\_2 & 11+15 & 66.67 / 69.70 & 81.17 / 91.68 & 30.51 / 27.43 & 52.79 / 58.26 \\
 & shift\_3 & 11+15 & 61.21 / 73.94 & 81.55 / 91.90 & 28.02 / 23.18 & 51.11 / 59.35 \\
 & underexposed & 11+15 & 54.55 / 74.55 & 85.37 / 91.23 & 17.98 / 21.47 & 40.29 / 50.15 \\
\addlinespace[2pt]
fruit\_jelly & regular & 5+15 & 86.67 / 82.67 & 95.45 / 96.22 & 43.28 / 35.77 & 77.33 / 77.45 \\
 & overexposed & 5+15 & 90.67 / 96.00 & 95.52 / 96.73 & 47.01 / 46.22 & 77.85 / 80.26 \\
 & shift\_1 & 5+15 & 77.33 / 57.33 & 94.32 / 94.84 & 37.70 / 23.90 & 74.24 / 71.43 \\
 & underexposed & 5+15 & 68.00 / 65.33 & 94.19 / 94.51 & 31.10 / 19.38 & 72.50 / 70.28 \\
\addlinespace[2pt]
rice & regular & 7+15 & 76.19 / 73.33 & 93.81 / 94.42 & 46.55 / 39.87 & 66.60 / 60.93 \\
 & overexposed & 7+15 & 75.24 / 66.67 & 93.72 / 94.34 & 44.70 / 38.00 & 66.32 / 59.89 \\
 & shift\_1 & 7+15 & 68.57 / 67.62 & 93.41 / 93.42 & 45.98 / 35.41 & 64.80 / 64.48 \\
 & shift\_2 & 7+15 & 70.48 / 79.05 & 95.12 / 93.51 & 44.65 / 34.90 & 71.03 / 64.05 \\
 & shift\_3 & 7+15 & 69.52 / 82.86 & 94.97 / 93.92 & 48.34 / 42.27 & 71.22 / 68.28 \\
 & underexposed & 7+15 & 73.33 / 71.43 & 93.53 / 94.16 & 46.80 / 40.56 & 66.33 / 60.33 \\
\addlinespace[2pt]
sheet\_metal & regular & 4+15 & 73.33 / 41.67 & 89.83 / 89.76 & 16.11 / 9.81 & 32.05 / 22.66 \\
 & overexposed & 4+15 & 73.33 / 43.33 & 89.69 / 89.42 & 15.95 / 9.72 & 31.56 / 22.04 \\
 & shift\_1 & 4+15 & 58.33 / 58.33 & 88.62 / 89.21 & 12.22 / 6.57 & 29.50 / 22.68 \\
 & shift\_2 & 4+15 & 60.00 / 65.00 & 88.38 / 89.06 & 11.69 / 6.21 & 29.52 / 22.42 \\
 & shift\_3 & 4+15 & 53.33 / 56.67 & 88.79 / 89.40 & 12.30 / 6.90 & 29.75 / 22.93 \\
 & underexposed & 4+15 & 75.00 / 46.67 & 89.89 / 89.91 & 16.20 / 9.31 & 32.54 / 22.88 \\
\addlinespace[2pt]
vial & regular & 5+15 & 88.00 / 82.67 & 90.25 / 88.27 & 70.77 / 65.95 & 90.80 / 89.35 \\
 & overexposed & 5+15 & 96.00 / 86.67 & 87.64 / 85.76 & 66.53 / 60.56 & 89.38 / 86.75 \\
 & shift\_1 & 5+15 & 94.67 / 94.67 & 92.24 / 89.88 & 65.01 / 54.65 & 90.40 / 85.84 \\
 & shift\_2 & 5+15 & 93.33 / 81.33 & 90.33 / 87.81 & 70.29 / 60.19 & 91.11 / 87.25 \\
 & shift\_3 & 5+15 & 90.67 / 78.67 & 91.89 / 90.11 & 67.23 / 60.27 & 90.64 / 88.20 \\
 & shift\_4 & 5+15 & 94.67 / 94.67 & 92.87 / 90.84 & 61.89 / 52.80 & 89.40 / 85.49 \\
 & underexposed & 5+15 & 94.67 / 52.00 & 84.86 / 72.89 & 59.64 / 46.47 & 84.02 / 71.95 \\
\addlinespace[2pt]
wallplugs & regular & 10+15 & 36.00 / 40.67 & 89.47 / 90.53 & 20.24 / 23.48 & 53.22 / 50.59 \\
 & overexposed & 10+15 & 34.67 / 41.33 & 89.41 / 90.47 & 19.76 / 23.28 & 52.90 / 50.36 \\
 & shift\_1 & 10+15 & 50.67 / 53.33 & 92.06 / 91.26 & 23.46 / 19.99 & 60.23 / 54.21 \\
 & shift\_2 & 10+15 & 37.33 / 27.33 & 86.57 / 80.29 & 17.70 / 11.46 & 45.04 / 36.33 \\
 & shift\_3 & 10+15 & 34.00 / 28.67 & 91.56 / 89.38 & 24.10 / 22.88 & 57.90 / 52.91 \\
 & underexposed & 10+15 & 37.33 / 42.67 & 89.51 / 90.61 & 20.15 / 23.33 & 51.94 / 50.69 \\
\addlinespace[2pt]
walnuts & regular & 10+15 & 92.67 / 94.67 & 97.45 / 97.26 & 60.05 / 63.99 & 85.85 / 86.39 \\
 & overexposed & 10+15 & 92.67 / 94.67 & 97.43 / 97.17 & 60.34 / 64.69 & 85.75 / 86.33 \\
 & shift\_1 & 10+15 & 55.33 / 56.67 & 94.98 / 93.21 & 33.17 / 22.29 & 72.44 / 68.14 \\
 & shift\_2 & 10+15 & 90.00 / 94.00 & 97.36 / 96.98 & 59.72 / 61.94 & 85.21 / 85.42 \\
 & shift\_3 & 10+15 & 59.33 / 70.67 & 95.56 / 94.09 & 37.83 / 28.67 & 75.46 / 71.95 \\
 & underexposed & 10+15 & 91.33 / 94.67 & 97.46 / 97.28 & 59.04 / 63.08 & 85.78 / 86.57 \\
\bottomrule\end{tabular}\end{table}
\end{document}